\documentclass[10pt,a4paper,twocolumn]{article}
\usepackage[T1]{fontenc}
\usepackage[utf8]{inputenc}
\usepackage{amsmath,amssymb}
\usepackage{txfonts}
\usepackage{graphicx}
\usepackage[a4paper,top=20mm,bottom=22mm,left=15mm,right=15mm,columnsep=6mm]{geometry}
\usepackage{multirow,threeparttable,gensymb}
\usepackage[authoryear,round]{natbib}
\usepackage{xcolor,url}
\usepackage[font=small,labelfont=bf]{caption}
\usepackage[colorlinks=true,linkcolor=blue,urlcolor=blue,citecolor=black]{hyperref}
\hypersetup{pdftitle={PoRef-MIL: Plug-and-Play Posterior Refinement for Multiple Instance Learning in QSM-Based Brain Disease Classification},pdfauthor={Ang Gao; Rui Jin; Yin Liu; Pengfei Rong; Feng Liu; G. Bruce Pike; Hongfu Sun; Yang Gao}}
\begin{document}
\twocolumn[{
\begin{center}
{\LARGE\bfseries PoRef-MIL: Plug-and-Play Posterior Refinement for Multiple Instance Learning in QSM-Based Brain Disease Classification\par}
\vspace{10pt}
{\normalsize Ang Gao\textsuperscript{1,\textdagger}, Rui Jin\textsuperscript{1,\textdagger}, Yin Liu\textsuperscript{2,*}, Pengfei Rong\textsuperscript{2},
Feng Liu\textsuperscript{3}, G. Bruce Pike\textsuperscript{4}, Hongfu Sun\textsuperscript{5}, Yang Gao\textsuperscript{1,*}\par}
\vspace{6pt}
{\footnotesize
\textsuperscript{1}School of Computer Science and Engineering, Central South University, Changsha, China\par
\textsuperscript{2}Department of Radiology, The Third Xiangya Hospital, Central South University, Changsha, China\par
\textsuperscript{3}School of Electrical Engineering and Computer Science, University of Queensland, Brisbane, Australia\par
\textsuperscript{4}Departments of Radiology and Clinical Neurosciences, Hotchkiss Brain Institute, University of Calgary, Calgary, Canada\par
\textsuperscript{5}School of Engineering, University of Newcastle, Newcastle, Australia\par
\vspace{4pt}
\textsuperscript{\textdagger}These two authors contributed equally to this work.\par
\textsuperscript{*}Correspondence to \href{mailto:liuyin99@foxmail.com}{Dr. Yin Liu} and \href{mailto:yang.gao@csu.edu.cn}{Dr. Yang Gao}.\par}
\end{center}
\vspace{4pt}
\begin{center}\textbf{Abstract}\end{center}
\begin{minipage}{\textwidth}
\small
Quantitative susceptibility mapping (QSM) is a novel quantitative MRI contrast for studying tissue susceptibility changes in brain disorders, including Parkinson's disease (PD) and multiple sclerosis (MS). However, small cohorts and the lack of local annotations make developing deep learning models for QSM-based diagnosis challenging. In this work, we propose PoRef-MIL, a weakly supervised multiple instance learning (MIL) framework combining iterative slice relabeling and network retraining with a plug-and-play posterior refinement module (PRM). During training, a 2D dual-branch diagnosis network and a subject-level classifier were alternately trained for five rounds. The classifier was used to update slice pseudo-labels for retraining the diagnosis network. The PRM was applied during inference to refine slice probabilities through bidirectional posterior inference and calculate entropy-derived confidence from the refined probabilities, which was then used in sliding-window scoring to obtain subject-level predictions. PoRef-MIL was evaluated on an in-house PD cohort of 287 participants and a public MS cohort of 150 participants. Compared with eight classification methods, PoRef-MIL achieved the highest mean areas under the receiver operating characteristic curve of $0.942$ and $0.872$, and accuracies of $0.926$ and $0.833$, for PD and MS classification, respectively. Ablation experiments were conducted to validate the effectiveness of iterative relabeling and PRM, and the plug-and-play feature of the proposed PRM was also evaluated using four state-of-the-art MIL models. Overall, PoRef-MIL offers a promising approach to QSM-based disease classification in small cohorts without manual slice annotations.
\par\medskip
\noindent\textbf{Keywords:} Quantitative susceptibility mapping (QSM); Plug-and-play posterior refinement module (PRM); Brain disease classification; \mbox{Multiple instance learning (MIL)}.
\end{minipage}
\vspace{16pt}
}]

\section{Introduction}
\label{sec:introduction}
Quantitative susceptibility mapping (QSM) is a novel MRI post-processing technique capable of calculating the spatial distribution of tissue magnetic susceptibility from phase measurements \citep{wang2015quantitative,liu2015quantitative,qsmconsensus2024}, which is sensitive to changes in paramagnetic iron and diamagnetic myelin and provides quantitative information to investigate tissue variations in a range of neurological disorders \citep{ayton2017cerebral,sun2018quantitative,chen2021quantitative}. For instance, in Parkinson's disease (PD), previous QSM studies have reported increased susceptibility in the substantia nigra, together with susceptibility alterations in other subcortical and cortical regions, highlighting the regional distribution of disease-related abnormalities \citep{langkammer2016quantitative,acostacabronero2017wholebrain,guan2022altered,guan2024qsm},  while in multiple sclerosis (MS), lesion susceptibility is reported to be influenced by both myelin loss and iron accumulation \citep{langkammer2013ms, wisnieff2015lesions,deh2018myelin,voon2024qsmms}. QSM has also demonstrated great potential in studying inflammatory diseases and characterizing differences in lesion tissue properties and their evolution \citep{kaunzner2019inflammation,rahmanzadeh2022remyelination,elkady2025longitudinalprl}.

Existing studies have developed several approaches to extracting disease-related patterns from 3D QSM images, including hybrid feature extraction \citep{xiao2019hybrid, zhang2022qsmrim}, radiomics-based classification \citep{li20193d, kang2022radiomics}, and multimodal machine learning that combines susceptibility and metabolic information \citep{yu2026midbrain, tian2026multimodal}. Some deep neural networks have also been adopted for PD classification \citep{wang2023pipeline} and for identifying chronic active MS lesions by combining lesion-level convolutional and radiomic features \citep{zhang2022qsmrim}. In addition to disease classification, QSM-based graph neural networks have been investigated for assessing postural abnormalities, rigidity, and arising-from-chair abnormalities in PD patients \citep{tang2023postural,tang2024causalqsm, tang2024rigidity}. Despite these promising applications, deep learning-based disease classification using QSM remains relatively underexplored.

Both limited training data and the granularity of available annotations hinder broader development of deep learning-based QSM disease classification methods. In medical imaging, deep models often need to be developed using relatively small patient cohorts, which provide only a small number of examples for inter-subject variability and disease-related imaging patterns \citep{cheplygina2019notsosupervised,jin2026lesssupervision}. In addition, subject-level diagnostic labels do not specify which slices or regions contain disease-related abnormalities \citep{liu2018landmarkmil,cheplygina2019notsosupervised,weisser2024weakly}. Although these diagnostic labels can supervise subject-level classification, they do not provide the local targets required for fully supervised slice-level learning, and therefore cannot provide slice- or region-level evidence. Simply assigning the subject's diagnosis to every slice will neglect the spatially heterogeneous distribution of abnormalities within a brain volume \citep{cheplygina2019notsosupervised,weisser2024weakly,jin2026lesssupervision}. These limitations motivate weakly supervised approaches that can learn disease-relevant local patterns from the subject-level labels.

Multiple instance learning (MIL) provides a suitable framework by representing each subject as a bag of image instances and aggregating local image information for subject-level classification without requiring explicit instance-level annotations \citep{carbonneau2018milreview,cheplygina2019notsosupervised,  vandervelden2022xai, jin2026lesssupervision}. This paradigm has been widely adopted for gigapixel whole-slide image (WSI) classification in computational pathology \citep{lu2021clam,shao2021transmil,zhang2022dtfdmil,cai2025attrimil, yan2024shapley}. It has also been applied to brain MRI classification through different strategies for representing and aggregating local image information \citep{liu2018landmarkmil,zhu2021dualmil,zhang2023smimil}. Previous studies have explored landmark-based instance construction \citep{liu2018landmarkmil}, dual-attention aggregation \citep{zhu2021dualmil}, multi-instance knowledge distillation \citep{guan2021mri}, anatomy-guided instance selection \citep{jiang2024anatomy}, and slice- or patch-based aggregation for subject- or examination-level diagnosis \citep{zhang2023smimil}. There are also some methods investigating attention-based pooling, inter-instance correlation modeling and feature distillation to improve instance representation and bag-level prediction \citep{ilse2018attentionmil,shao2021transmil,zhang2022dtfdmil}. In QSM analysis, MIL-based graph networks have also incorporated intervention and counterfactual strategies to reduce the influence of irrelevant or spuriously relevant features \citep{tang2023postural,tang2024causalqsm}.

However, several challenges remain when applying MIL to QSM-based brain disease classification. For example, accurate subject-level classification does not necessarily imply reliable instance-level predictions \citep{vandervelden2022xai, weisser2024weakly,cai2025attrimil}, which is particularly relevant to 3D QSM images, where disease-related susceptibility changes may be
concentrated in specific regions rather than distributed uniformly throughout the brain \citep{acostacabronero2017wholebrain,guan2024qsm,rahmanzadeh2022remyelination}. For interpreting QSM-based predictions, explicit slice-level labels accompanied by prediction probabilities and uncertainty information would allow the distribution of model-identified abnormalities to be examined in greater detail. Furthermore, reducing annotation requirements in MIL does not totally eliminate the limitations imposed by small training cohorts \citep{liu2018landmarkmil,weisser2024weakly}, which may result in a decrease in performance due to insufficient inter-subject diversity. Most previous MIL methods chose to improve classification performance through novel designs in more efficient feature learning \citep{liu2018landmarkmil, zhang2022dtfdmil}, attention mechanisms \citep{ilse2018attentionmil,lu2021clam}, or instance aggregation \citep{shao2021transmil} within the trained model.

\begin{figure*}[t!]
\centering
\includegraphics[width=0.95\textwidth]{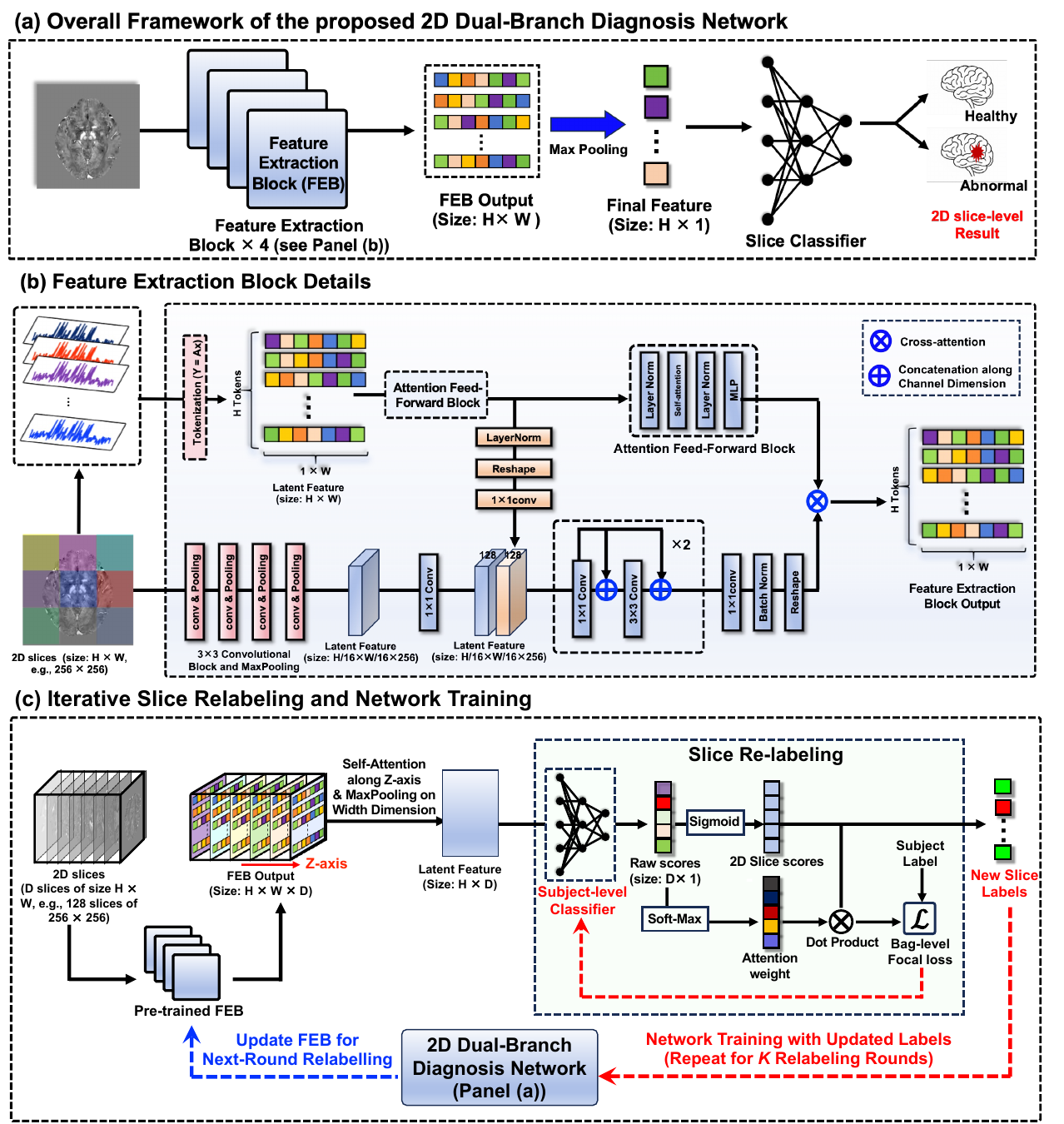}
\caption{Overview of the 2D dual-branch diagnosis network and its iterative training in PoRef-MIL. (a) In the 2D diagnosis network, each 2D QSM slice is processed by four feature extraction blocks (FEBs), followed by max pooling and slice-level classification. (b) Each FEB block fuses the features from the transformer-based and convolutional branches through cross-attention. (c) Latent features extracted from all 2D slices within a 3D volume are processed by self-attention along the slice direction, and slice scores are then aggregated using attention weights to compute a bag-level focal loss under subject-level supervision. Pseudo-labels derived from these scores are used to retrain the diagnosis network and update the FEB for the next relabeling round. This cycle is repeated for $K$ rounds.}
\label{fig:training-framework}
\end{figure*}

In this study, we propose PoRef-MIL, a multiple instance learning framework with a plug-and-play Posterior Refinement Module (PRM) for QSM-based brain disease classification. PoRef-MIL treats each 3D QSM volume as a sequence of consecutive 2D slices and uses a 2D dual-branch diagnosis network to predict slice-level abnormality probabilities. The iterative training strategy alternates between training a subject-level classifier to generate slice pseudo-labels (relabeling procedure) and retraining the diagnosis network using these updated labels. The PRM refines the ordered slice-level predictions through bidirectional posterior inference, followed by sliding-window scoring weighted by entropy-derived confidence to obtain final subject-level predictions. PoRef-MIL was evaluated on an in-house PD QSM dataset and a publicly available MS QSM dataset \citep{fiscone2024dataset} through comparisons with eight state-of-the-art methods. Slice-level gradient-weighted class activation mapping (Grad-CAM) heatmaps \citep{grad-cam} were generated to examine the spatial regions contributing to model predictions, together with slice abnormality probability distributions, refined slice states, and confidence curves. Ablation experiments were conducted to assess the effectiveness of iterative relabeling and the PRM. The module was further applied to four state-of-the-art MIL-based classifiers \citep{zhu2021dualmil,cai2025attrimil,yan2024shapley,shao2021transmil}, with the pretrained backbones kept frozen, to demonstrate its plug-and-play capability. Experiments combining QSM with complementary MRI contrasts, including T1-weighted (T1w) and fluid-attenuated inversion recovery (FLAIR) images, were also conducted to evaluate the framework's applicability to multimodal classification. Finally, age conditioning using feature-wise linear modulation (FiLM) \citep{perez2018film} was also evaluated to examine the influence of incorporating age information on classification performance.

\section{Method}
\label{sec:method}
\subsection{Overview of the Proposed PoRef-MIL Framework}
\label{subsec:overview}
The proposed PoRef-MIL framework performs subject-level brain disease classification from 3D QSM volumes under weak supervision, using only subject-level diagnostic labels. It is also able to produce slice-level abnormality probability and confidence profiles, and generates slice-level pseudo-labels during training. Following many previous MIL formulations \citep{ilse2018attentionmil,liu2018landmarkmil,carbonneau2018milreview,shao2021transmil,yan2024shapley,cai2025attrimil}, each subject is represented as a bag of image instances in this work, corresponding to consecutive 2D slices extracted from a 3D QSM volume.

Specifically, the 2D dual-branch diagnosis network (DBD-Net) combines a vision transformer (ViT) branch \citep{dosovitskiy2021vit} with a convolutional neural network (CNN) branch to extract image features and predict an abnormality probability for each slice (Fig. \ref{fig:training-framework}(a)). DBD-Net is trained using an iterative strategy that alternates between slice relabeling (i.e., training a subject-level classifier to generate slice pseudo-labels) and network retraining (i.e., retraining the slice-level DBD-Net using these updated labels), as detailed in Section \ref{subsec:training}.

\begin{figure*}[t!]
\centering
\includegraphics[width=0.9\textwidth]{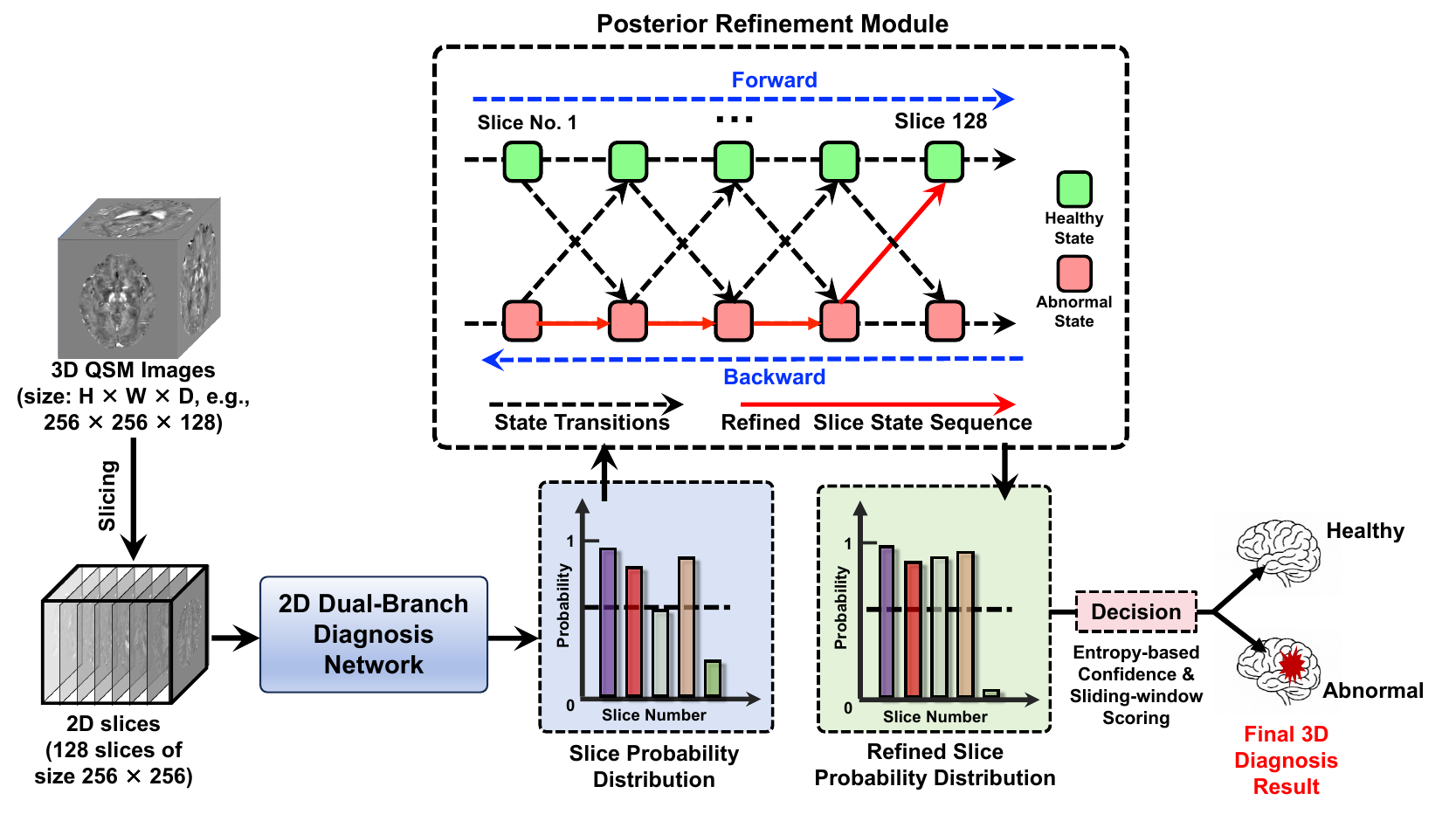}
\caption{Posterior Refinement Module (PRM) in the proposed PoRef-MIL inference framework. The 3D QSM volumes are treated as a consecutive series of 2D slices, which are fed into the pretrained 2D dual-branch diagnosis network to obtain initial slice probabilities. The slice probabilities and states are then refined using a bidirectional posterior refinement module. After the refinement, entropy-derived confidence is calculated from these probabilities and used in sliding-window scoring to determine the final subject-level predictions. }
\label{fig:PRM}
\end{figure*}

During inference, a plug-and-play posterior refinement module (PRM) refines the ordered slice-level abnormality probabilities produced by the pretrained DBD-Net. Entropy-derived confidence is then calculated from the refined probabilities and incorporated into a sliding-window scoring process to obtain the final subject-level prediction, as illustrated in Fig. \ref{fig:PRM} and detailed in Section \ref{subsec:inference}.

\subsection{The 2D Dual-Branch Diagnosis Network and Its Iterative Training}
\label{subsec:training}

As illustrated in Fig. \ref{fig:training-framework}(a), the proposed DBD-Net consists of four sequential feature extraction blocks (FEBs) and an MLP-based slice classifier. The features extracted by the FEBs are reduced to a 256-dimensional vector via max pooling and then passed to an MLP classifier with successive hidden-layer widths of 512, 1024, 512, 256, 128, and 2, followed by a single output unit for slice-level abnormality prediction. The structure of each FEB is detailed in Fig. \ref{fig:training-framework}(b), where each FEB is composed of a transformer branch for modeling relationships among image tokens and a convolutional branch for extracting local spatial features. In the transformer branch, tokenized features are processed by two consecutive attention feed-forward blocks following the transformer architecture \citep{vaswani2017attention, dosovitskiy2021vit}, each containing layer normalization, multi-head self-attention, and a two-layer feed-forward network that expands the embedding dimension from 256 to 512 and projects it back to 256. In parallel, the convolutional branch initially applies four convolutional stages, each consisting of a $3\times3$ convolution, batch normalization, ReLU activation, and max pooling. The resulting latent features are then processed by a $1\times1$ convolutional layer to adjust the number of channels to 128. Next, a feature bridge consisting of layer normalization, reshaping, and a $1\times1$ convolution is introduced to transfer intermediate transformer features to the convolutional branch, via channel-wise concatenation. The combined features are then fed into two consecutive residual convolutional blocks, each comprising a $1\times1$ convolution and a $3\times3$ convolution. Finally, features from the convolutional and transformer branches are fused using cross-attention to produce the FEB output.

In this work, the proposed DBD-Net was trained using an iterative strategy that alternates between slice relabeling and network retraining to obtain slice-level supervision using only subject-level diagnostic labels, as illustrated in Fig. \ref{fig:training-framework}(c).

Initially, the diagnostic label of each training subject was assigned to all slices within the corresponding QSM volume for an initial warm-up training. These initial slice labels were used to train DBD-Net by minimizing a binary focal loss \citep{lin2017focal} calculated for individual slices, as in traditional supervised learning.

After the initial warm-up training, the weights of the four pretrained FEBs were transferred to a slice relabeling module. Specifically, for each 3D QSM volume (data size: $H \times W \times D$), features extracted from all $D$ slices were arranged in their original order and processed by self-attention along the slice direction to incorporate information across all slices. The resulting latent feature representation of each slice was then passed through max pooling (along the second dimension) and an MLP to produce a raw score $s_i$, where $i=1,2,\ldots,D$ denotes the slice index within the 3D volume. This relabeling classifier used the same layer configuration as the DBD-Net slice classifier, but its parameters were learned separately. The same raw scores were passed through the sigmoid and softmax operations to obtain the slice abnormality probabilities and corresponding attention weights, respectively:
\begin{equation}
p_i = \sigma(s_i) = \frac{1}{1+\exp(-s_i)}, \quad
a_i = \frac{\exp(s_i)}{\sum_{j=1}^{D}\exp(s_j)},
\label{eq:relabel-slice-scores}
\end{equation}
where $p_i \in [0, 1]$ denotes the slice abnormality probability for slice $i$, with $0$ representing a healthy slice and $1$ denoting a potentially abnormal slice, and $a_i$ denotes its attention weight. The softmax operation normalizes the scores across the slices of the same subject, such that $\sum_{i=1}^{D}a_i=1$.

Similar to previous attention-weighted  aggregations \citep{ilse2018attentionmil}, the slice abnormality probabilities $p_i$ were combined to obtain a subject-level probability ($P_{subject}$):
\begin{equation}
P_{\mathrm{subject}} = \sum_{i=1}^{D} a_i p_i,
\label{eq:relabel-bag-probability}
\end{equation}
and this subject-level probability was then used to calculate the subject-level focal loss, based on the original subject-level diagnostic label ($Y\in\{0,1\}$, with $0$ and $1$ denoting healthy and diseased subjects, respectively), as shown below:
\begin{equation}
\mathcal{L}_{\mathrm{subject}}
= -\alpha_t(1-P_t)^{\gamma}\log(P_t),
\label{eq:relabel-bag-loss}
\end{equation}
where $P_t=YP_{\mathrm{subject}}+(1-Y)(1-P_{\mathrm{subject}})$, $\alpha_t=\alpha Y+(1-\alpha)(1-Y)$ is the class weight, and $\gamma$ is the focusing parameter. In this work, $\alpha$ and $\gamma$ were empirically set as $0.5$ and $2$, respectively. This loss was back-propagated to optimize the relabeling classifier (i.e., subject-level classifier in Fig. \ref{fig:training-framework}(c)) for up to 300 epochs, while the pretrained FEB weights remained frozen during relabeling.

After this subject-level classifier was trained, the proposed relabeling module in Fig. \ref{fig:training-framework}(c) was subsequently used to predict slice abnormality probabilities ({$P=\{p_i\}$, $i=1,2,\ldots,D$}), and new slice pseudo-labels ($\tilde{Y}=\{\tilde{y}_i\}$, where $i$ is the slice index) were generated correspondingly based on the following Eq. \ref{eq:relabel-pseudo-labels}, allowing the slice-level retraining and network refinement without manual slice annotations:
\begin{equation}
\tilde{y}_i =
\begin{cases}
1, & p_i \geq 0.5\\
0, & p_i < 0.5
\end{cases},
\label{eq:relabel-pseudo-labels}
\end{equation}

Then, each pseudo-label paired with its corresponding original QSM slice was used for next-round DBD-Net re-training, and this relabeling module was used only during network training procedures. Overall, this network training and slice relabeling framework was repeated for $K$ rounds, and in each round, both the proposed DBD-Net (Fig. \ref{fig:training-framework}(a)) and the subject-level classifier (Fig. \ref{fig:training-framework}(c)) were reinitialized and trained from scratch.

\subsection{Posterior Refinement Module and Inference in PoRef-MIL}
\label{subsec:inference}

As shown in Fig. \ref{fig:PRM}, during inference, each 3D QSM volume (data size: $H\times W\times D$) was divided into $D$ consecutive slices, which were fed into the pretrained DBD-Net to obtain the initial slice abnormality probabilities ($P=\{p_i\}$, where $i=1,2,\ldots,D$). Rather than directly assigning a healthy or abnormal status to each slice by thresholding its initial probability, the proposed PRM refined these probabilities by incorporating information from both preceding and following slices.

In this work, each slice was assumed to be associated with two possible states, with 0 and 1 denoting healthy and abnormal, respectively. The slice abnormality probabilities predicted by DBD-Net were used to construct a slice observation evidence matrix ($E=[e_i(s)]_{s,i}$, where $s\in\{0,1\}$ and $i=1,2,\ldots,D$) for these two states:
\begin{equation}
e_i(0)=1-p_i, \quad e_i(1)=p_i,
\label{eq:prm-observation}
\end{equation}
where $e_i(0)$ and $e_i(1)$ represent the evidence supporting the healthy and abnormal states of slice $i$, respectively.

Then, a symmetric $2\times2$ transition matrix $A$ was introduced to model the state dependence between consecutive slices, whose element $A_{rs}$ denoted the probability that the current slice is in state $s$ given that the preceding slice is in state $r$, with $\sum_{s=0}^{1}A_{rs}=1$ for $r\in\{0,1\}$. For example, $A_{00}$ and $A_{01}$ represent the probabilities that the current slice is healthy or abnormal, respectively, given that the preceding slice is healthy. In this work, the transition matrix was empirically set as
\begin{equation}
A =
\begin{bmatrix}
0.9 & 0.1\\
0.1 & 0.9
\end{bmatrix},
\label{eq:prm-transition}
\end{equation}
favoring the same state in consecutive slices while still allowing transitions between healthy and abnormal states.

Next, forward and backward recursions for a two-state hidden Markov model (HMM) with fixed transition probabilities \citep{rabiner1989tutorial} were performed along the entire slice sequence to refine the initial slice abnormality probabilities using information from both directions. The forward quantity $\alpha_i(s)$ represents the accumulated support for state $s$ at slice $i$ based on the observation evidence from slices $1$ to $i$, while the backward quantity $\beta_i(s)$ denotes the support provided by the subsequent slices, from slice $D$ to $i+1$, conditional on slice $i$ being in state $s$. Specifically, $\alpha_i(s)$ was calculated as follows:
\begin{equation}
\alpha_i(s)=
\frac{
e_i(s)\displaystyle\sum_{r=0}^{1}\alpha_{i-1}(r)A_{rs}
}{
\displaystyle\sum_{u=0}^{1}
e_i(u)\sum_{r=0}^{1}\alpha_{i-1}(r)A_{ru}
},
\label{eq:prm-forward}
\end{equation}
where $s\in\{0,1\}$, $i=1,2\ldots,D$. The numerator part combines evidence propagated from the previous slice with the current slice evidence. The denominator sums these results over the two possible states, ensuring that $\alpha_i(0)+\alpha_i(1)=1$. $r$ denotes the preceding state, and $s$ denotes the current state.

Similarly, the backward quantity $\beta_i(s)$ was calculated from the end of the slice sequence using the corresponding recursion:
\begin{equation}
\beta_i(s)=
\frac{
\displaystyle\sum_{t=0}^{1}
A_{st}e_{i+1}(t)\beta_{i+1}(t)
}{
\displaystyle\sum_{u=0}^{1}\sum_{t=0}^{1}
A_{ut}e_{i+1}(t)\beta_{i+1}(t)
},
\label{eq:prm-backward}
\end{equation}
where $i=D-1,D-2,\ldots,1$, and $t$ denotes the state of the next slice. Similarly, the numerator combines the support propagated from the next slice, while the denominator ensures that $\beta_i(0)+\beta_i(1)=1$.

The forward and backward recursions were initialized as
\begin{equation}
\alpha_0(s)=0.5,\quad
\beta_D(s)=0.5,\quad s\in\{0,1\},
\label{eq:prm-initialization}
\end{equation}
where $\alpha_0(s)$ specifies equal initial support for the two possible states (healthy or abnormal) before the first slice. Since no slices follow slice $D$, the backward boundary $\beta_D(s)$ assigns equal weights to both states, representing the absence of further observation evidence.

After the bidirectional recursions, the refined abnormality probability $q_i$ of slice $i$ was calculated by combining the forward and backward support:
\begin{equation}
q_i=
\frac{\alpha_i(1)\beta_i(1)}
{\displaystyle\sum_{u=0}^{1}\alpha_i(u)\beta_i(u)},
\label{eq:prm-posterior}
\end{equation}
where $1-q_i$ denotes the corresponding probability of the healthy state; the numerator combines the support for the abnormal state from both directions, while the denominator normalizes the combined support over the healthy and abnormal states. Therefore, $q_i$ incorporates observation evidence from the entire slice sequence through the state transitions specified by $A$.

Following posterior probability refinement, a binary Shannon entropy $H_i$ and the corresponding confidence $C_i$ were also calculated for each slice:
\begin{equation}
\begin{cases}
H_i = -q_i\log_2(q_i)
       -(1-q_i)\log_2(1-q_i)\\
C_i = 1-H_i
\end{cases},
\label{eq:prm-confidence}
\end{equation}
where $H_i,C_i\in[0,1]$, with $0\log_2 0$ defined as $0$. When $q_i$ approaches 0.5, the two states have similar probabilities, resulting in high entropy and low confidence. Conversely, probabilities approaching either 0 or 1 will yield low entropy and high confidence. Overall, $q_i$ describes the predicted abnormality of a slice, whereas $C_i$ describes the confidence of this binary prediction.

Finally, to obtain a subject-level score, a sliding window of length $L$ (empirically set to 10 in this work) was traversed along the whole probability sequence with a stride of 1. For a window starting at slice $j$, the refined probabilities were averaged using the corresponding confidence values as weights:
\begin{equation}
S_{\mathrm{subject}} = \max_{j\in\mathcal{J}} W_j = \max_{j\in\mathcal{J}}
\frac{\displaystyle\sum_{i=j}^{j+L-1}C_iq_i}
{\displaystyle\sum_{i=j}^{j+L-1}C_i},
\label{eq:prm-window}
\end{equation}
where $\mathcal{J}=\{1,\ldots,D-L+1\}$ denotes the set of available window starting positions. The normalization by the confidence sum allowed slices with higher confidence to contribute more to the decision. The final subject-level classification ($\hat{Y}_{subject}$) was obtained by the following simple thresholding:
\begin{equation}
\hat{Y}_{subject} =
\begin{cases}
1\quad\text{(Diseased)}, & S_{\mathrm{subject}}> 0.5\\
0\quad\text{(Healthy)}, & S_{\mathrm{subject}}\leq 0.5
\end{cases}.
\label{eq:prm-subject-classification}
\end{equation}

\section{Experiments}
\label{sec:experiments}

\subsection{Dataset Acquisition and Partitioning}
\label{subsec:datasets}

\begin{table}[t!]
\centering
\begingroup
\setlength{\tabcolsep}{1.6pt}
\renewcommand{\arraystretch}{1.15}
\begin{threeparttable}
\caption{Demographic characteristics of the PD and MS cohorts}
\small
\label{tab:cohort-demographics}
\begin{tabular}{@{}lcccc@{}}
\hline\noalign{\vskip 3pt}
\multirow{2}{*}{\textbf{Characteristics}} & \multicolumn{2}{c}{\textbf{PD dataset}} & \multicolumn{2}{c}{\textbf{MS dataset}} \\
\cline{2-3}\cline{4-5}
 & \textbf{HC} & \textbf{PD} & \textbf{HC} & \textbf{MS} \\
\noalign{\vskip 2pt}\hline\noalign{\vskip 2pt}
Participant Counts: & 131 & 156 & 50 & 100 \\
Age (years)\tnote{1} & $59.4 \pm 8.34$ & $60.7 \pm 8.08$ & $57.1 \pm 16.7$ & $47.6 \pm 11.4$ \\
Gender (M/F)\tnote{2} & 55 / 76 & 81 / 75 & 19 / 31 & 42 / 58 \\
\noalign{\vskip 3pt}\hline
\end{tabular}
\par\vspace{3pt}
\begin{tablenotes}[flushleft]
\footnotesize\raggedright
\item[] HC, healthy controls; PD, Parkinson's disease; MS, multiple sclerosis.
\item[1] Age is reported as mean $\pm$ standard deviation.
\item[2] Male/female values indicate participant counts.
\end{tablenotes}
\end{threeparttable}
\endgroup
\end{table}

\begin{table*}[t!]
\centering
\begingroup
\setlength{\tabcolsep}{2.5pt}
\setlength{\arrayrulewidth}{0.6pt}
\renewcommand{\arraystretch}{1.15}
\begin{threeparttable}
\caption{Comparison of the Proposed PoRef-MIL with Eight State-of-the-Art Methods on PD and MS Datasets}
\label{tab:method-comparison}

\begin{tabular}{@{}lccccc@{}}
\hline\noalign{\vskip 3pt}
\textbf{Method} & \textbf{AUC} & \textbf{ACC} & \textbf{SEN} & \textbf{SPE} & \textbf{F1 Score} \\
\noalign{\vskip 2pt}\hline\noalign{\vskip 2pt}

\multicolumn{6}{@{}l}{\textbf{PD dataset}} \\
\noalign{\vskip 1pt}

\textbf{AG-SE-ResNeXt50 \citep{wang2023pipeline}}
 & $0.893 \pm 0.002^{\boldsymbol{**}}$
 & $0.846 \pm 0.015^{\boldsymbol{**}}$
 & $0.858 \pm 0.018^{\boldsymbol{**}}$
 & $0.831 \pm 0.021^{\boldsymbol{**}}$
 & $0.858 \pm 0.014^{\boldsymbol{**}}$ \\

\textbf{MMRN \citep{han2023multi}}
 & $0.899 \pm 0.002^{\boldsymbol{**}}$
 & $0.860 \pm 0.012^{\boldsymbol{**}}$
 & $0.832 \pm 0.014^{\boldsymbol{**}}$
 & $0.892 \pm 0.032^{\boldsymbol{*}}$
 & $0.866 \pm 0.010^{\boldsymbol{**}}$ \\

\textbf{CAAE + CRAT \citep{cobbinah2022multicenter}}
 & $0.903 \pm 0.003^{\boldsymbol{**}}$
 & $0.870 \pm 0.020^{\boldsymbol{**}}$
 & $0.858 \pm 0.018^{\boldsymbol{**}}$
 & $0.885 \pm 0.027^{\boldsymbol{*}}$
 & $0.878 \pm 0.019^{\boldsymbol{**}}$ \\

\textbf{DenseFormer-MoE \citep{ding2025denseformer}}
 & $0.906 \pm 0.001^{\boldsymbol{**}}$
 & $0.856 \pm 0.015^{\boldsymbol{**}}$
 & $0.813 \pm 0.014^{\boldsymbol{**}}$
 & $\underline{0.908 \pm 0.021}$
 & $0.860 \pm 0.014^{\boldsymbol{**}}$ \\

\noalign{\vskip 3pt}

\textbf{PMIL \citep{yan2024shapley}}
 & $0.890 \pm 0.013^{\boldsymbol{**}}$
 & $0.853 \pm 0.010^{\boldsymbol{**}}$
 & $0.865 \pm 0.014^{\boldsymbol{**}}$
 & $0.838 \pm 0.017^{\boldsymbol{**}}$
 & $0.865 \pm 0.009^{\boldsymbol{**}}$ \\

\textbf{DA-MIDL \citep{zhu2021dualmil}}
 & $0.914 \pm 0.002^{\boldsymbol{**}}$
 & $0.867 \pm 0.010^{\boldsymbol{**}}$
 & $0.865 \pm 0.014^{\boldsymbol{**}}$
 & $0.869 \pm 0.021^{\boldsymbol{**}}$
 & $0.876 \pm 0.009^{\boldsymbol{**}}$ \\

\textbf{AttriMIL \citep{cai2025attrimil}}
 & $0.912 \pm 0.002^{\boldsymbol{**}}$
 & $0.874 \pm 0.008^{\boldsymbol{**}}$
 & $0.884 \pm 0.018^{\boldsymbol{**}}$
 & $0.862 \pm 0.021^{\boldsymbol{**}}$
 & $0.884 \pm 0.007^{\boldsymbol{**}}$ \\

\textbf{TransMIL \citep{shao2021transmil}}
 & $\underline{0.927 \pm 0.001}^{\boldsymbol{**}}$
 & $\underline{0.898 \pm 0.008}^{\boldsymbol{**}}$
 & $\underline{0.890 \pm 0.018}^{\boldsymbol{*}}$
 & $\underline{0.908 \pm 0.021}$
 & $\underline{0.905 \pm 0.008}^{\boldsymbol{**}}$ \\

\textbf{PoRef-MIL (proposed)}
 & $\boldsymbol{0.942 \pm 0.002}$
 & $\boldsymbol{0.926 \pm 0.008}$
 & $\boldsymbol{0.923 \pm 0.018}$
 & $\boldsymbol{0.931 \pm 0.017}$
 & $\boldsymbol{0.932 \pm 0.008}$ \\

\noalign{\vskip 2pt}\hline\noalign{\vskip 2pt}

\multicolumn{6}{@{}l}{\textbf{MS dataset \citep{fiscone2024dataset}}} \\
\noalign{\vskip 1pt}

\textbf{AG-SE-ResNeXt50 \citep{wang2023pipeline}}
 & $0.792 \pm 0.051^{\boldsymbol{*}}$
 & $0.753 \pm 0.061$
 & $0.780 \pm 0.057$
 & $0.700 \pm 0.071$
 & $0.808 \pm 0.049$ \\

\textbf{MMRN \citep{han2023multi}}
 & $0.838 \pm 0.029$
 & $0.800 \pm 0.053$
 & $0.820 \pm 0.057$
 & $0.760 \pm 0.055$
 & $0.845 \pm 0.043$ \\

\textbf{CAAE + CRAT \citep{cobbinah2022multicenter}}
 & $0.812 \pm 0.029^{\boldsymbol{*}}$
 & $0.767 \pm 0.033^{\boldsymbol{*}}$
 & $0.780 \pm 0.027^{\boldsymbol{*}}$
 & $0.740 \pm 0.055$
 & $0.817 \pm 0.026^{\boldsymbol{*}}$ \\

\textbf{DenseFormer-MoE \citep{ding2025denseformer}}
 & $0.783 \pm 0.034^{\boldsymbol{**}}$
 & $0.740 \pm 0.049^{\boldsymbol{*}}$
 & $0.760 \pm 0.042^{\boldsymbol{*}}$
 & $0.700 \pm 0.071$
 & $0.796 \pm 0.040^{\boldsymbol{*}}$ \\

\noalign{\vskip 3pt}

\textbf{PMIL \citep{yan2024shapley}}
 & $0.790 \pm 0.021^{\boldsymbol{**}}$
 & $0.740 \pm 0.028^{\boldsymbol{**}}$
 & $0.830 \pm 0.027$
 & $0.560 \pm 0.055^{\boldsymbol{**}}$
 & $0.810 \pm 0.021^{\boldsymbol{*}}$ \\

\textbf{DA-MIDL \citep{zhu2021dualmil}}
 & $0.802 \pm 0.033^{\boldsymbol{*}}$
 & $0.773 \pm 0.043$
 & $0.790 \pm 0.042$
 & $0.740 \pm 0.055$
 & $0.823 \pm 0.035$ \\

\textbf{AttriMIL \citep{cai2025attrimil}}
 & $0.820 \pm 0.010^{\boldsymbol{*}}$
 & $0.780 \pm 0.018$
 & $\boldsymbol{0.870 \pm 0.027}$
 & $0.600 \pm 0.000^{\boldsymbol{**}}$
 & $0.840 \pm 0.015$ \\

\textbf{TransMIL \citep{shao2021transmil}}
 & $\underline{0.851 \pm 0.041}$
 & $\underline{0.813 \pm 0.056}$
 & $0.830 \pm 0.045$
 & $\underline{0.780 \pm 0.084}$
 & $\underline{0.856 \pm 0.043}$ \\

\textbf{PoRef-MIL (proposed)}
 & $\boldsymbol{0.872 \pm 0.034}$
 & $\boldsymbol{0.833 \pm 0.053}$
 & $\underline{0.850 \pm 0.050}$
 & $\boldsymbol{0.800 \pm 0.071}$
 & $\boldsymbol{0.871 \pm 0.042}$ \\

\noalign{\vskip 3pt}\hline
\end{tabular}

\begin{tablenotes}[flushleft]
\footnotesize\raggedright
\item[] Note: The metrics are presented as mean $\pm$ standard deviation across five models trained using five-fold cross-validation within the development sets. All five models were evaluated on the same fixed test set for each dataset. Bold and underlined values indicate the highest and second-highest means, respectively.
\item[] $t$-tests were conducted to compare each method with the proposed PoRef-MIL, with $^{\boldsymbol{*}}$ and $^{\boldsymbol{**}}$ indicating $P<0.05$ and $P<0.01$, respectively.
\end{tablenotes}

\end{threeparttable}
\endgroup
\end{table*}

The proposed PoRef-MIL was evaluated separately on an in-house PD dataset and a publicly available MS dataset \citep{fiscone2024dataset}.  The PD dataset was collected at the Third Xiangya Hospital, Central South University, and included a total of 287 participants, comprising 156 patients with PD and 131 healthy controls (HCs). The public MS dataset included 150 participants, comprising 100 patients with MS and 50 HCs. The demographic characteristics of both datasets are summarized in Table \ref{tab:cohort-demographics}. These datasets were used for two separate binary classification tasks, distinguishing patients with PD or MS from the corresponding HCs (i.e., HC vs. PD and HC vs. MS, respectively).

Institutional ethics board approvals have been obtained for the in-house MRI brain data used in this work. For the PD dataset, all QSM scans were acquired on a Philips 3T scanner (Ingenia Elition X; Philips Medical Systems, Best, the Netherlands) with a 32-channel receiver array head coil and the following sequence and parameters: 3D multi-echo GRE with 10 unipolar echoes, FOV = $224 \times 224 \times 128 \mathrm{mm}^{3}$, flip angle = 20$\degree$, BW = 855 Hz/pixel, TE\textsubscript{1} / $\Delta$ TE / TR = 4.2 / 4.1 / 43.4 ms, spatial resolution = 0.8 mm $\times$ 0.8 mm $\times$ 1 mm, CS-SENSE acceleration factor = 3, and scan time = 4.6 min. The recently developed single-step iQSM+ method \citep{GAO2024103160} was used for QSM reconstruction from phase images, after brain mask segmentation using BET \citep{smith2002fast} based on the magnitude images. For each subject, the whole-brain mean susceptibility was subtracted from the reconstructed QSM map, setting the mean susceptibility within the brain to zero. In addition, sagittal T1-weighted (T1w) structural images were also acquired using a three-dimensional turbo field echo (3D-TFE) sequence with the following parameters: TR/TE = 8.3/3.8 ms, flip angle = 8$\degree$, turbo factor = 240, FOV = $240 \times 240 \times 170 \mathrm{mm}^{3}$, voxel size =  0.94 mm $\times$ 0.94 mm $\times$ 1 mm, acquisition matrix = 256 $\times$ 256 $\times$ 170, compressed sensing acceleration factor = 3, and acquisition time = 3 min 5 s. These T1w images were used together with QSM images in the multimodal classification experiments for the HC vs. PD task.

The public MS dataset also included T1w and fluid-attenuated inversion recovery (FLAIR) images in addition to the QSM images. Details of the acquisition protocols are described in the original dataset publication \citep{fiscone2024dataset}. QSM images were used for the main classification experiments, while T1w and FLAIR images were combined with QSM in the additional multimodal MS classification experiments.

For each dataset, a fixed test set was held out before model training. The PD test set included 31 patients with PD and 26 HCs, while the MS test set included 20 patients with MS and 10 HCs. The remaining participants formed the development set for each dataset. All partitions were defined at the subject level, with all slices from the same subject assigned to the same subset.

Five-fold cross-validation was performed within each development set during network training. In each fold, four subsets were used for training and the remaining subset was used for validation and model selection (early stopping). The complete iterative training procedure was performed separately for each fold, yielding five models for each dataset. All five models were then evaluated individually on the fixed test set, which was excluded from model training and selection.

\begin{table*}[t!]
\centering
\begingroup
\setlength{\tabcolsep}{4pt}
\setlength{\arrayrulewidth}{0.6pt}
\renewcommand{\arraystretch}{1.15}
\begin{threeparttable}
\caption{Plug-and-play Effectiveness of the Proposed Posterior Refinement Module (PRM) Across the Five MIL Models}
\label{tab:prm-ablation}
\begin{tabular}{@{}lcccccc@{}}
\hline\noalign{\vskip 3pt}
\multirow{2}{*}{\textbf{Model Configurations}} & \multicolumn{3}{c}{\textbf{PD dataset}} & \multicolumn{3}{c}{\textbf{MS dataset \citep{fiscone2024dataset}}} \\
\cline{2-4}\cline{5-7}
 & \textbf{AUC} & \textbf{ACC} & \textbf{F1 Score} & \textbf{AUC} & \textbf{ACC} & \textbf{F1 Score} \\
\noalign{\vskip 2pt}\hline\noalign{\vskip 2pt}
\textbf{AttriMIL w/o PRM} & $0.912 \pm 0.002$ & $0.874 \pm 0.008$ & $0.884 \pm 0.007$ & $0.820 \pm 0.010$ & $0.780 \pm 0.018$ & $0.840 \pm 0.015$ \\
\textbf{AttriMIL + PRM} & $0.920 \pm 0.003$ & $0.888 \pm 0.010$ & $0.898 \pm 0.008$ & $0.844 \pm 0.012$ & $0.807 \pm 0.028$ & $0.859 \pm 0.021$ \\
\noalign{\vskip 2pt}\hline\noalign{\vskip 2pt}
\textbf{DA-MIDL w/o PRM} & $0.914 \pm 0.002$ & $0.867 \pm 0.010$ & $0.876 \pm 0.009$ & $0.802 \pm 0.033$ & $0.773 \pm 0.043$ & $0.823 \pm 0.035$ \\
\textbf{DA-MIDL + PRM} & $0.924 \pm 0.003$ & $0.891 \pm 0.015$ & $0.899 \pm 0.014$ & $0.834 \pm 0.015$ & $0.807 \pm 0.028$ & $0.850 \pm 0.022$ \\
\noalign{\vskip 2pt}\hline\noalign{\vskip 2pt}
\textbf{PMIL w/o PRM} & $0.890 \pm 0.013$ & $0.853 \pm 0.010$ & $0.865 \pm 0.009$ & $0.790 \pm 0.021$ & $0.740 \pm 0.028$ & $0.810 \pm 0.021$ \\
\textbf{PMIL + PRM} & $0.912 \pm 0.003$ & $0.884 \pm 0.016$ & $0.893 \pm 0.015$ & $0.830 \pm 0.013$ & $0.813 \pm 0.030$ & $0.857 \pm 0.023$ \\
\noalign{\vskip 2pt}\hline\noalign{\vskip 2pt}
\textbf{TransMIL w/o PRM} & $0.927 \pm 0.001$ & $0.898 \pm 0.008$ & $0.905 \pm 0.008$ & $0.851 \pm 0.041$ & $0.813 \pm 0.056$ & $0.856 \pm 0.043$ \\
\textbf{TransMIL + PRM} & $0.934 \pm 0.005$ & $0.909 \pm 0.015$ & $0.916 \pm 0.013$ & $0.868 \pm 0.029$ & $0.827 \pm 0.043$ & $0.868 \pm 0.034$ \\
\noalign{\vskip 2pt}\hline\noalign{\vskip 2pt}

\textbf{PoRef-MIL w/o PRM\textsuperscript{1}}
 & $0.925 \pm 0.005$ & $0.898 \pm 0.015$ & $0.906 \pm 0.014$
 & $0.851 \pm 0.033$ & $0.800 \pm 0.033$ & $0.845 \pm 0.029$ \\
\textbf{PoRef-MIL w/o PRM\textsuperscript{2}} & $0.938 \pm 0.006$ & $0.909 \pm 0.015$ & $0.918 \pm 0.013$ & $0.865 \pm 0.032$ & $0.820 \pm 0.038$ & $0.863 \pm 0.030$ \\
\textbf{PoRef-MIL (proposed)} & $\boldsymbol{0.942 \pm 0.002}$ & $\boldsymbol{0.926 \pm 0.008}$ & $\boldsymbol{0.932 \pm 0.008}$ & $\boldsymbol{0.872 \pm 0.034}$ & $\boldsymbol{0.833 \pm 0.053}$ & $\boldsymbol{0.871 \pm 0.042}$ \\
\noalign{\vskip 3pt}\hline
\end{tabular}

\begin{tablenotes}[flushleft]
\footnotesize\raggedright
\item[] \textbf{PoRef-MIL w/o PRM\textsuperscript{1}}
uses the FEBs and subject-level classifier from the final relabeling
stage ($K=5$) to directly predict subject-level
probabilities without PRM.
\textbf{PoRef-MIL w/o PRM\textsuperscript{2}}
uses the final DBD-Net ($K=5$) to predict slice-level
probabilities, which are then averaged within a sliding window
of 10 slices with a stride of one slice. The maximum
window-averaged probability is used as the subject-level
score, with values greater than $0.5$ classified as positive
and the remaining values classified as HC.
\end{tablenotes}

\end{threeparttable}
\endgroup
\end{table*}

\subsection{Network Training Details}
\label{subsec:training-details}
PoRef-MIL was implemented in PyTorch 2.9.1 and trained on an NVIDIA RTX A6000 GPU. The iterative training procedure described in Section \ref{subsec:training} was performed. Specifically, DBD-Net was first trained using the subject-level diagnostic labels assigned to individual slices in the initial warm-up training, with a batch size of 8 subjects (using all slices of each subject). Then, slice relabeling was conducted and a subject-level classifier was trained using one complete subject per batch, comprising 128 axial slices for the HC vs. PD task and 96 axial slices for the HC vs. MS task. After subject-level classifier training, new slice pseudo-labels were generated, enabling the retraining of the slice-level DBD-Net with the original QSM slices and these updated labels. These two procedures alternated with each other for $K = 5$ rounds in this work. Both the slice-level DBD-Net and the subject-level classifier (during the relabeling procedure) were optimized using the Adam optimizer with a weight decay of $1\times10^{-5}$ and binary focal loss \citep{lin2017focal}, with $\alpha=0.5$ and $\gamma=2$.

The initial learning rates for the five specific training rounds were set to $1\times10^{-4}$, $5\times10^{-5}$, $3\times10^{-5}$, $1\times10^{-5}$, and $5\times10^{-6}$, respectively. Each training stage was conducted for a maximum of 300 epochs, with early stopping applied. The ReduceLROnPlateau scheduler was used to monitor subject-level validation performance using the area under the receiver operating characteristic curve (AUC), with a learning rate reduction factor of 0.5, a patience of 10 epochs, and a minimum learning rate of $1\times10^{-7}$. A new checkpoint was saved when the validation AUC exceeded the previously recorded best value by more than $1\times10^{-4}$. Training stopped after 50 consecutive epochs without such an improvement. The complete training of all five models took approximately 24 hours in total, including both DBD-Net training and the intervening relabeling-classifier training.

\begin{figure}[b!]
\centering
\includegraphics[width=\columnwidth]{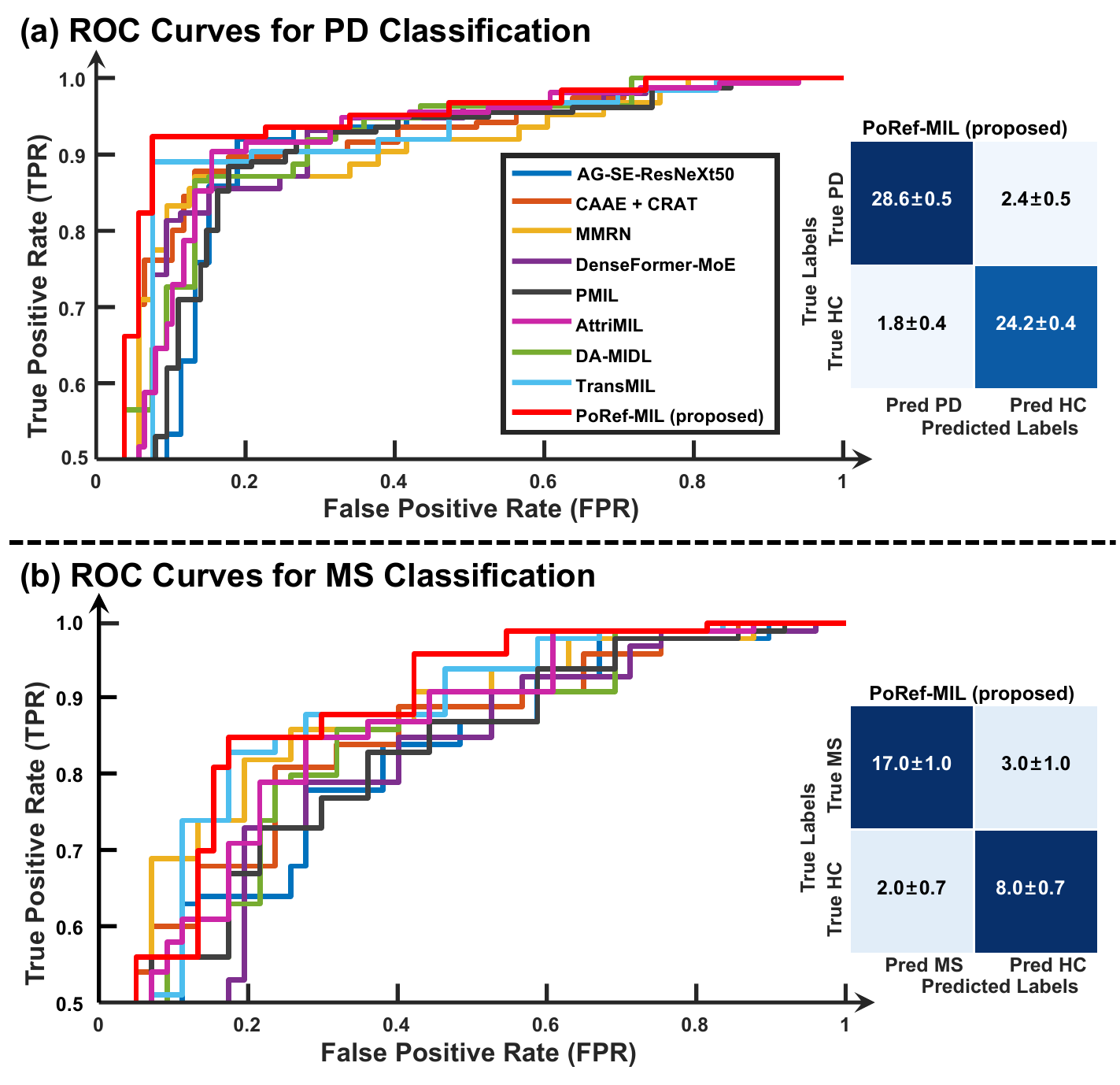}
\caption{Comparison of receiver operating characteristic (ROC) curves of the proposed PoRef-MIL with the other methods on PD (a) and  MS (b) test sets, with corresponding confusion matrices of PoRef-MIL results (mean $\pm$ standard deviation across the five trained models) shown on the right. }
\label{fig:roc-confusion}
\end{figure}

\begin{figure*}[t!]
\centering
\includegraphics[width=0.9\textwidth]{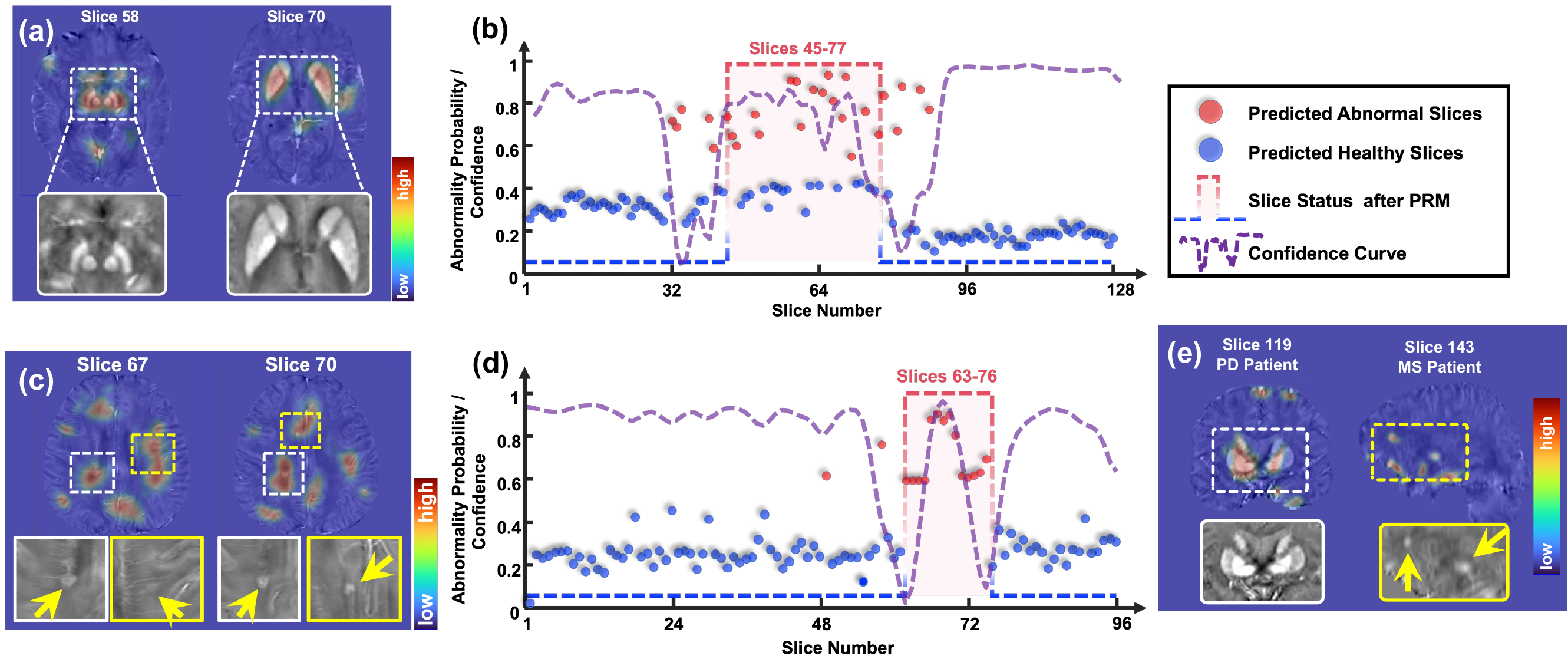}
\caption{Grad-CAM maps and slice-level predictions for representative PD (a, b) and MS (c, d) subjects. (a) and (c) show Grad-CAM maps overlaid on the original QSM slices, with warmer colors indicating stronger contributions. (b) and (d) show the initial slice-level abnormality probability distributions, with red and blue dots indicating slices predicted as abnormal and healthy, respectively. The red and blue dashed step lines represent the corresponding slice states after PRM, with shaded intervals marking slices 45 to 77 in the PD subject and slices 63 to 76 in the MS subject. Purple dashed curves indicate entropy-derived confidence calculated from the refined slice-level probabilities. (e) shows additional coronal and sagittal Grad-CAM heatmaps for a PD subject and an MS subject, respectively, generated using separately trained models. }
\label{fig:slice-interpretation}
\end{figure*}

\subsection{Performance Evaluation and Metrics}
\label{subsec:evaluation}
The proposed PoRef-MIL was compared with four subject-level classification methods, including an attention-gated squeeze-and-excitation ResNeXt50 model used for PD diagnosis (AG-SE-ResNeXt50) \citep{wang2023pipeline}, CAAE+CRAT \citep{cobbinah2022multicenter}, a multi-template meta-information regularized network (MMRN) \citep{han2023multi}, a dense Transformer foundation model with a mixture of experts (DenseFormer-MoE) \citep{ding2025denseformer}, and four MIL methods, i.e., dual attention multi-instance deep learning (DA-MIDL) \citep{zhu2021dualmil}, AttriMIL \citep{cai2025attrimil}, PMIL \citep{yan2024shapley}, and Transformer-based correlated multiple instance learning (TransMIL) \citep{shao2021transmil}, using AUC, accuracy (ACC), sensitivity (SEN), specificity (SPE), and F1 score (with patients treated as the positive class and HCs as the negative class). Each metric was calculated separately for the five trained models of each task, and the results were reported as mean $\pm$ standard deviation (SD). Receiver operating characteristic (ROC) curves were also generated by pooling the subject-level prediction scores and corresponding diagnostic labels from the five models evaluated on the same fixed test set for each task.

Ablation experiments were also conducted to evaluate the contribution of the proposed PRM based on both the in-house PD and public MS datasets. The PRM was also applied to DA-MIDL, AttriMIL, PMIL, and TransMIL to evaluate its plug-and-play capability and its contribution to classification performance across different models. Furthermore, the effectiveness of the iterative relabeling strategy was examined by varying the iterative training rounds from $K=0$ to $K=5$, with $K=0$ representing a model trained without the slice relabeling.

All PoRef-MIL results in this work were obtained using models trained on axial slices, except for the additional orthogonal qualitative heatmaps generated using two models trained separately on coronal slices from the PD cohort and sagittal slices from the MS cohort. Grad-CAM heatmaps \citep{grad-cam} were illustrated for representative PD and MS subjects, together with slice abnormality probability distributions, the slice states after PRM, and the entropy-derived confidence curves. The numbers of model-predicted abnormal slices were also compared between the HC and patient groups in both datasets using $t$-tests.

Additional experiments were also conducted to evaluate the applicability of PoRef-MIL to different MRI modalities. For the PD task, two modalities (T1w and QSM) were evaluated, while for the MS task, T1w, QSM, and FLAIR images were evaluated. Finally, the influence of incorporating age information on classification performance was investigated by comparing models with and without feature-wise linear modulation (FiLM)-based age conditioning \citep{perez2018film}.

\section{Results}
\label{sec:results}

\subsection{Performance Comparisons with State-of-the-Art Methods}
\label{subsec:classification-performance}
Table \ref{tab:method-comparison} reports the classification performance comparisons of the proposed PoRef-MIL with several state-of-the-art methods on the in-house PD and public MS test datasets. Overall, PoRef-MIL achieved the highest mean AUC, ACC, SPE, and F1 score on both classification tasks. Specifically, on the HC vs. PD dataset, PoRef-MIL achieved the highest mean AUC of 0.942, ACC of 0.926, and F1 score of 0.932, which were 0.015, 0.028, and 0.027 significantly ($P<0.01$) higher than those of TransMIL (the second-best results), respectively. In addition, PoRef-MIL achieved the highest mean AUC of $0.872 \pm 0.034$, ACC of $0.833 \pm 0.053$, and F1 score of $0.871 \pm 0.042$ among all compared methods for the HC vs. MS task, exceeding those of TransMIL by 0.021, 0.020, and 0.015, respectively.

Figure \ref{fig:roc-confusion} compares the ROC curves of PoRef-MIL with AG-SE-ResNeXt50, CAAE + CRAT, MMRN, DenseFormer-MoE, DA-MIDL, AttriMIL, PMIL, and TransMIL. PoRef-MIL generally showed higher true positive rates in the low false positive rate region for both classification tasks. In addition, the confusion matrices of the proposed PoRef-MIL are also reported, with subject counts presented as mean $\pm$ SD across the five trained models. Overall, PoRef-MIL correctly classified on average 28.6 of the 31 patients with PD and 24.2 of the 26 HCs in the PD test set, and 17.0 of the 20 patients with MS and 8.0 of the 10 HCs in the MS test set.

\begin{table}[t!]
\centering
\begingroup
\setlength{\tabcolsep}{8pt}
\setlength{\arrayrulewidth}{0.6pt}
\renewcommand{\arraystretch}{1.15}
\begin{threeparttable}
\caption{Classification Performance with Conventional MIL and Different Iterative Slice Relabeling Rounds}
\label{tab:relabeling-rounds}
\begin{tabular*}{\linewidth}{@{\extracolsep{\fill}}lccc@{}}
\hline\noalign{\vskip 3pt}
\textbf{\textit{K}} & \textbf{AUC} & \textbf{ACC} & \textbf{F1 Score} \\
\noalign{\vskip 2pt}\hline\noalign{\vskip 2pt}
\multicolumn{4}{@{}l}{\textbf{PD dataset}} \\
\textbf{--}\tnote{a} & $0.912 \pm 0.002$ & $0.895 \pm 0.012$ & $0.903 \pm 0.012$ \\
\textbf{0} & $0.667 \pm 0.027$ & $0.639 \pm 0.016$ & $0.673 \pm 0.012$ \\
\textbf{1} & $0.913 \pm 0.006$ & $0.877 \pm 0.018$ & $0.888 \pm 0.015$ \\
\textbf{2} & $0.924 \pm 0.005$ & $0.898 \pm 0.015$ & $0.906 \pm 0.014$ \\
\textbf{3} & $0.932 \pm 0.005$ & $0.909 \pm 0.015$ & $0.916 \pm 0.013$ \\
\textbf{4} & $0.938 \pm 0.004$ & $0.916 \pm 0.008$ & $0.922 \pm 0.008$ \\
\textbf{5} & $\boldsymbol{0.942 \pm 0.002}$ & $\boldsymbol{0.926 \pm 0.008}$ & $\boldsymbol{0.932 \pm 0.008}$ \\
\noalign{\vskip 2pt}\hline\noalign{\vskip 2pt}
\multicolumn{4}{@{}l}{\textbf{MS dataset}} \\
\textbf{--}\tnote{a} & $0.838 \pm 0.003$ & $0.793 \pm 0.015$ & $0.839 \pm 0.011$ \\
\textbf{0} & $0.623 \pm 0.033$ & $0.713 \pm 0.051$ & $0.816 \pm 0.031$ \\
\textbf{1} & $0.810 \pm 0.032$ & $0.760 \pm 0.049$ & $0.814 \pm 0.039$ \\
\textbf{2} & $0.831 \pm 0.032$ & $0.787 \pm 0.045$ & $0.835 \pm 0.036$ \\
\textbf{3} & $0.854 \pm 0.033$ & $0.813 \pm 0.045$ & $0.857 \pm 0.035$ \\
\textbf{4} & $0.866 \pm 0.032$ & $0.827 \pm 0.043$ & $0.867 \pm 0.035$ \\
\textbf{5} & $\boldsymbol{0.872 \pm 0.034}$ & $\boldsymbol{0.833 \pm 0.053}$ & $\boldsymbol{0.871 \pm 0.042}$ \\
\noalign{\vskip 3pt}\hline
\end{tabular*}
\begin{tablenotes}[flushleft]
\footnotesize\raggedright
\item[] $K$: number of slice relabeling rounds; $K=0$ denotes models trained using slice-level loss without slice relabeling.
\item[a] Conventional MIL models trained end-to-end using subject-level
loss; $K$ is not applicable.
\end{tablenotes}
\end{threeparttable}
\endgroup
\end{table}

\begin{table}[b!]
\centering
\begingroup
\small
\setlength{\tabcolsep}{2pt}
\setlength{\arrayrulewidth}{0.6pt}
\renewcommand{\arraystretch}{1.1}
\begin{threeparttable}
\caption{Single-modality and multimodal classification with PoRef-MIL}
\label{tab:multimodal-classification}
\begin{tabular*}{\linewidth}{@{\extracolsep{\fill}}lccc@{}}
\hline\noalign{\vskip 3pt}
\textbf{Inputs} & \textbf{AUC} & \textbf{ACC} & \textbf{F1 Score} \\
\noalign{\vskip 2pt}\hline\noalign{\vskip 2pt}
\multicolumn{4}{@{}l}{\textbf{PD dataset}} \\
\textbf{T1w} & $0.908 \pm 0.012$ & $0.898 \pm 0.015$ & $0.906 \pm 0.014$ \\
\textbf{QSM} & $\underline{0.942 \pm 0.002}$ & $\underline{0.926 \pm 0.008}$ & $\underline{0.932 \pm 0.008}$ \\
\textbf{T1w+QSM} & $\boldsymbol{0.964 \pm 0.004}$ & $\boldsymbol{0.944 \pm 0.015}$ & $\boldsymbol{0.949 \pm 0.013}$ \\
\noalign{\vskip 2pt}\hline\noalign{\vskip 2pt}
\multicolumn{4}{@{}l}{\textbf{MS dataset}} \\
\textbf{T1w} & $0.844 \pm 0.030$ & $0.800 \pm 0.053$ & $0.848 \pm 0.040$ \\
\textbf{QSM} & $0.872 \pm 0.034$ & $0.833 \pm 0.053$ & $0.871 \pm 0.042$ \\
\textbf{FLAIR} & $\underline{0.877 \pm 0.028}$ & $\underline{0.840 \pm 0.049}$ & $\underline{0.877 \pm 0.038}$ \\
\textbf{T1w+QSM+FLAIR} & $\boldsymbol{0.887 \pm 0.028}$ & $\boldsymbol{0.860 \pm 0.043}$ & $\boldsymbol{0.894 \pm 0.032}$ \\
\noalign{\vskip 3pt}\hline
\end{tabular*}
\begin{tablenotes}[flushleft]
\footnotesize\raggedright
\item[] Bold and underlined values indicate the highest and second-highest means, respectively.
\end{tablenotes}
\end{threeparttable}
\endgroup
\end{table}

\subsection{Ablation Studies of the Posterior Refinement Module and Iterative Relabeling}
\label{subsec:ablation-results}
Table \ref{tab:prm-ablation} compares the classification performance of the complete PoRef-MIL with two prediction configurations without PRM on both datasets. On the PD dataset, incorporating the PRM achieved the best mean AUC, ACC, and F1 score of 0.942, 0.926, and 0.932, respectively, while the corresponding values were 0.872, 0.833, and 0.871 on the MS dataset.

In addition, the plug-and-play feature of PRM was also validated using the other four MIL models (keeping the pretrained  network backbones frozen). For example, as shown in Table \ref{tab:prm-ablation}, the TransMIL model combined with the plug-and-play PRM block achieved average increases of 0.012 in AUC, 0.0125 in ACC, and 0.0115 in F1 score across the two datasets compared with the original TransMIL model.

The influences of iterative relabeling and training rounds (i.e., $K$) are investigated in Table \ref{tab:relabeling-rounds}. The mean AUC, ACC, and F1 score consistently increased with more relabeling and retraining rounds (from $K=1$ to $K=5$, where $K=0$ denotes a model trained without the slice relabeling) on both datasets. Overall, all models trained with iterative relabeling ($K\ge1$) demonstrated dramatically better AUC and ACC results than the initial model ($K=0$). The first relabeling round increased the mean AUC from 0.667 to 0.913 on the PD dataset. As $K$ increased from 1 to 5, AUC, ACC, and F1 score gradually increased from 0.913, 0.877, and 0.888 to 0.942, 0.926, and 0.932, respectively, on the HC vs. PD dataset. In addition, models trained with different numbers of relabeling rounds were compared with a conventional end-to-end MIL model trained using subject-level loss without iterative slice relabeling. PoRef-MIL with $K\ge3$ achieved higher mean AUC, ACC, and F1 scores than conventional MIL on both datasets.

\begin{figure}[t!]
\centering
\includegraphics[width=0.95\columnwidth]{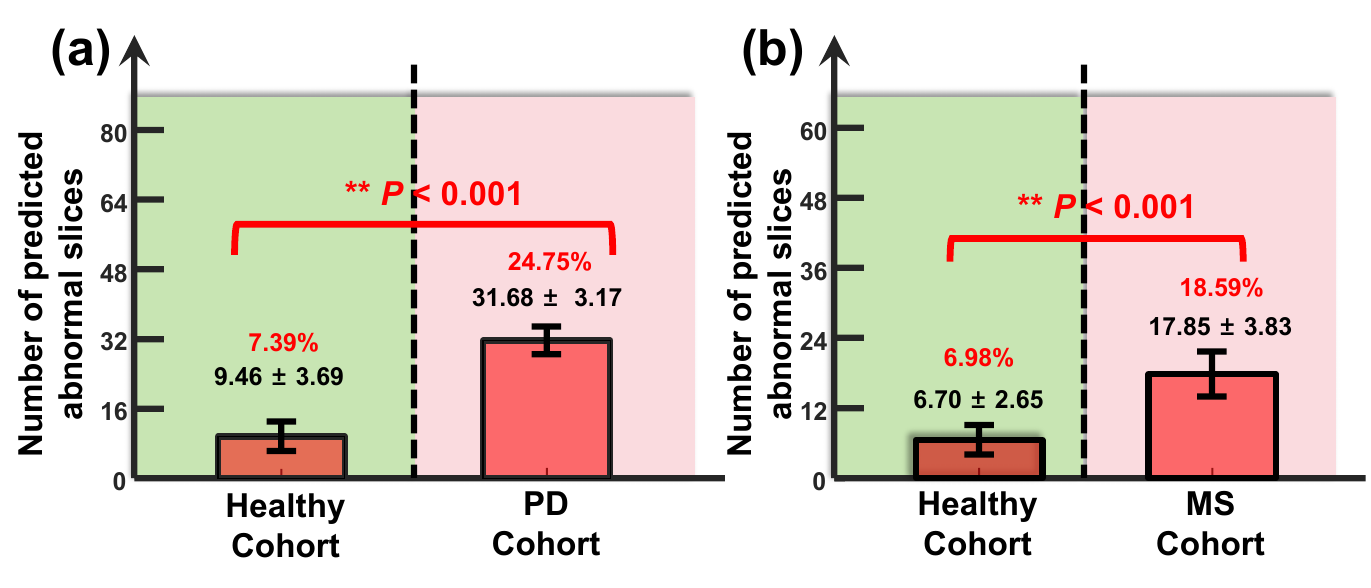}
\caption{Comparison of the numbers of model-predicted abnormal slices between healthy controls (HCs) and patients in the (a) PD and (b) MS test sets. Percentages indicate the mean proportion of predicted abnormal slices among all analyzed slices per subject (128 slices for PD and 96 slices for MS).}
\label{fig:slice-counts}
\end{figure}

\subsection{Slice-Level Predictions and Model Interpretability}
\label{subsec:interpretability}
Figure \ref{fig:slice-interpretation} illustrates the Grad-CAM heatmaps and slice-level prediction distributions for two representative PD and MS subjects. For the PD subject, stronger contributions were observed in deep gray matter regions on axial slices 58 and 70 (zoomed-in regions in Fig. \ref{fig:slice-interpretation}(a)) and coronal slice 119 in Fig. \ref{fig:slice-interpretation}(e). For the MS subject, several regions with stronger responses overlapped with the lesions visible in the QSM images, as indicated by the yellow arrows in the enlarged images of axial slices 67 and 70 (Fig. \ref{fig:slice-interpretation}(c)) and sagittal slice 143 in Fig. \ref{fig:slice-interpretation}(e).

The initial slice abnormality probabilities (red and blue dots) predicted by the DBD-Net, slice states after PRM (red and blue dashed step lines), and entropy-derived confidence (purple lines) are also plotted in Fig. \ref{fig:slice-interpretation}(b, d). After the PRM, slices 45 to 77 in the PD subject and slices 63 to 76 in the MS subject were classified as abnormal, while the remaining slices were classified as healthy. It is also noticed that some slices outside these intervals, although assigned high initial abnormality probabilities by the pretrained network, were still assigned healthy states after refinement. The confidence derived from the refined posterior probabilities also varied across the slice sequence. Overall, high confidence was observed for both predicted healthy and abnormal slices, with local decreases around several transitions between the two states.

Figure \ref{fig:slice-counts} compares the numbers of model-predicted abnormal slices between the HC and patient groups in both test sets. In both datasets, the patient groups showed significantly more predicted abnormal slices than the corresponding HC groups ($P<0.001$). Patients with PD had $31.68 \pm 3.17$ predicted abnormal slices per subject, compared with $9.46 \pm 3.69$ in the HC group. For the MS dataset, the corresponding numbers were $17.85 \pm 3.83$ in patients with MS and $6.70 \pm 2.65$ in HCs. In both datasets, the mean proportion of predicted abnormal slices was below 25\% in both the HC and patient groups, indicating that most slices were classified as healthy.

\subsection{Multimodal Classification and Age Conditioning}
\label{subsec:additional-results}
The classification performance of PoRef-MIL using single- and multi-modal MRI images is reported in Table \ref{tab:multimodal-classification}. In the PD dataset, combining T1w and QSM images achieved the highest mean AUC, ACC, and F1 score of 0.964, 0.944, and 0.949, respectively, compared with 0.942, 0.926, and 0.932 using QSM alone. For the HC vs. MS task, T1w, QSM, and FLAIR images together also achieved the best mean AUC, ACC, and F1 score of 0.887, 0.860, and 0.894, respectively.

Given the higher mean age of HCs than that of patients in the MS cohort, FiLM-based age conditioning was also evaluated, with the results reported in Table \ref{tab:age-conditioning}. Compared with the reference model using QSM images without additional age input, incorporating age information resulted in slightly lower mean AUC, ACC, and F1 scores on both datasets. For example, the mean AUC decreased from 0.942 to 0.931 on the PD dataset and from 0.872 to 0.860 on the MS dataset.

\begin{table}[t!]
\centering
\begingroup
\setlength{\tabcolsep}{2pt}
\setlength{\arrayrulewidth}{0.6pt}
\renewcommand{\arraystretch}{1.15}
\begin{threeparttable}
\caption{Evaluation of FiLM-based Age Conditioning in PoRef-MIL}
\label{tab:age-conditioning}
\begin{tabular*}{\linewidth}{@{\extracolsep{\fill}}lccc@{}}
\hline\noalign{\vskip 3pt}
\textbf{Age Input} & \textbf{AUC} & \textbf{ACC} & \textbf{F1 Score} \\
\noalign{\vskip 2pt}\hline\noalign{\vskip 2pt}
\multicolumn{4}{@{}l}{\textbf{PD dataset}} \\
\textbf{w/ FiLM} & $0.931 \pm 0.006$ & $0.916 \pm 0.008$ & $0.921 \pm 0.008$ \\
\textbf{w/o Age} & $\boldsymbol{0.942 \pm 0.002}$ & $\boldsymbol{0.926 \pm 0.008}$ & $\boldsymbol{0.932 \pm 0.008}$ \\
\noalign{\vskip 2pt}\hline\noalign{\vskip 2pt}
\multicolumn{4}{@{}l}{\textbf{MS dataset}} \\
\textbf{w/ FiLM} & $0.860 \pm 0.032$ & $0.813 \pm 0.061$ & $0.854 \pm 0.049$ \\
\textbf{w/o Age} & $\boldsymbol{0.872 \pm 0.034}$ & $\boldsymbol{0.833 \pm 0.053}$ & $\boldsymbol{0.871 \pm 0.042}$ \\
\noalign{\vskip 3pt}\hline
\end{tabular*}
\begin{tablenotes}[flushleft]
\footnotesize\raggedright
\item[] Age was embedded using a Transformer-style positional encoding and incorporated into image features through feature-wise linear modulation (FiLM) \citep{perez2018film}. The reference model uses QSM images without additional age input.
\end{tablenotes}
\end{threeparttable}
\endgroup
\end{table}

\section{Discussion and Conclusion}
\label{sec:discussion}
In this work, we proposed PoRef-MIL, a weakly supervised framework for QSM-based brain disease classification using only subject-level diagnostic labels. The framework combines iterative slice relabeling during network training with a plug-and-play posterior refinement module (PRM) during inference. Comprehensive experiments were conducted on an in-house PD dataset and a publicly available MS dataset \citep{fiscone2024dataset} to evaluate the proposed framework, including comparisons with several state-of-the-art subject-level classification and MIL methods. The results suggest that PoRef-MIL is a promising approach to disease classification in relatively small QSM cohorts without requiring manual slice-level annotations.

PoRef-MIL reformulates each 3D QSM volume as a sequence of consecutive 2D slices, allowing individual slices within the same subject to receive different predictions. The iterative relabeling strategy updates the slice-level supervision during network training, while the PRM incorporates information across the ordered slice sequence to refine predictions during inference. The ablation studies supported the contributions of both iterative relabeling and the PRM to the final classification performance. Both cohorts used in this work contained fewer than 300 participants, and the public MS dataset \citep{fiscone2024dataset} included only 150 participants. The performance obtained in these cohorts suggests that the proposed framework can learn disease-related imaging patterns from relatively small QSM datasets without requiring manual slice annotations.

According to the results in Fig. \ref{fig:slice-counts}, slices predicted by PoRef-MIL as healthy predominated in both the HC and patient groups, although patients had significantly more slices predicted as abnormal than HCs. This distribution is consistent with the motivation for allowing different slices within the same subject to have different training labels. When the diagnostic label of a patient is directly assigned to every slice, slices containing little disease-related information also receive the disease label, which might be problematic. By generating slice pseudo-labels, the iterative relabeling procedure may reduce the mismatch between inherited diagnostic labels and local image patterns. The lower AUC and ACC obtained without relabeling ($K=0$), together with the improvements after introducing relabeling, also support the value of updating slice-level supervision during network training.

Previous MIL studies have tried to improve classification performance through different strategies, including learned attention pooling \citep{ilse2018attentionmil}, clustering-constrained feature learning \citep{lu2021clam}, modeling of inter-instance correlations \citep{shao2021transmil}, and instance attribute constraints \citep{cai2025attrimil}. These developments improve how image information is represented and combined within the trained model. However, predictions from a trained network may still benefit from additional refinement before determining the final subject-level diagnosis, particularly when training cohorts are small and local annotations are unavailable. In this work, the network outputs were therefore treated as initial estimates of slice abnormality, and the PRM incorporates evidence from preceding and following slices to refine these estimates, while entropy-derived confidence determines their contributions to subsequent subject-level scoring. This additional inference procedure complements network training by using the information from the entire slice sequence to reassess local predictions before producing the final classification.

The PRM also has a plug-and-play property because it operates on ordered slice-level probabilities and does not depend on the internal feature representation of DBD-Net. For classifiers that already provide these probabilities, the PRM can be directly applied without further optimization of the network weights. For methods that only provide subject-level predictions, a slice-level classification head can be trained while keeping the pretrained backbone fixed before applying the PRM. In this work, the PRM was applied to DA-MIDL \citep{zhu2021dualmil}, AttriMIL \citep{cai2025attrimil}, PMIL \citep{yan2024shapley}, and TransMIL \citep{shao2021transmil}, with improvements in classification performance across these models.

In addition, PoRef-MIL provides slice abnormality probabilities, refined slice states, and entropy-derived confidence curves alongside the subject-level prediction. These outputs allow the distribution of model-predicted abnormalities and the associated confidence to be examined across the whole volume. In the representative MS examples, regions with stronger Grad-CAM responses overlapped well with lesions visible in the QSM images, providing qualitative support for the correspondence between model responses and local imaging abnormalities. The multimodal experiments further showed that complementary MRI contrasts could be incorporated through input-channel concatenation while retaining the same overall training and inference strategy.

However, there are still several limitations in this work. First, the alternating training of DBD-Net and the relabeling classifier increases the overall training time, while the number of relabeling rounds was also selected empirically, and $K=5$ may not be optimal for other datasets. Second, the age difference between HCs and patients in the MS cohort remains a potential source of confounding. Although FiLM-based age conditioning was evaluated, this experiment does not establish that the classification results were totally independent of age-related imaging differences. Third, the lack of slice-level reference annotations prevented direct assessment of the accuracy of the generated pseudo-labels and refined slice predictions. Finally, the evaluation was limited to two relatively small cohorts. Future studies involving different populations and acquisition settings are needed to better evaluate the generalizability of the proposed framework.

\section*{Data and Code Availability Statements}
Data are available on request due to privacy/ethical restrictions. Source codes and trained networks are available at: \url{https://github.com/YangGaoUQ/PoRef-MIL-QSM}.

\section*{Acknowledgments}
YG acknowledges that this work was supported by the National Natural Science Foundation of China under Grant No.	62301616, and the Natural Science Foundation of Hunan under Grant No. 2024JJ6530. HS acknowledges support from the Australian Research Council (DE210101297). GBP acknowledges support from Canadian Institutes of Health Research (FDN-143290).

\bibliographystyle{model2-names}
\bibliography{PoRefMIL}

\end{document}